\documentclass[letterpaper,10pt,conference]{ieeeconf}
\IEEEoverridecommandlockouts
\usepackage{amsmath}
\usepackage{amssymb}
\usepackage{booktabs}
\usepackage{cite}
\usepackage{cuted}
\usepackage{graphicx}
\usepackage{multirow}
\usepackage{tabularx}
\usepackage[table]{xcolor}
\usepackage[normalem]{ulem}
\usepackage[hidelinks]{hyperref}

\newcommand{\authorrefmarkl}[1]{\textsuperscript{#1}}

\makeatletter
\let\IEEEoriginalmakecaption\@makecaption
\long\def\@makecaption#1#2{%
    \ifx\@captype\@IEEEtablestring
        \begin{center}\footnotesize #1:\quad\normalfont #2\end{center}%
        \@IEEEtablecaptionsepspace
    \else
        \IEEEoriginalmakecaption{#1}{#2}%
    \fi}
\makeatother

\newcolumntype{Y}{>{\centering\arraybackslash}X}

\newcommand{\modelname}{Skel-WAM}

\title{\LARGE \bf
Skel-WAM: A Hand-Skeleton-Conditioned World Action Model for Human-to-Robot Manipulation Transfer
  \vspace{-5pt}
}

\author{
  \authorblockN{
    \textbf{Zetao Cai}\authorrefmarkl{1,2}\hspace{1.5em}
    \textbf{Yaping Li}\authorrefmarkl{1}\hspace{1.5em}
    \textbf{Yiqun Wang}\authorrefmarkl{2}\hspace{1.5em}
    \textbf{Xinyu Zhan}\authorrefmarkl{2}\hspace{1.5em}
    \textbf{Yuyin Yang}\authorrefmarkl{2}\\
    \textbf{Haoxiang Ma}\authorrefmarkl{2}\hspace{1.5em}
    \textbf{Kailin Li}\authorrefmarkl{2}\hspace{1.5em}
    \textbf{Tao Lu}\authorrefmarkl{2}\hspace{1.5em}
    \textbf{Jiangmiao Pang}\authorrefmarkl{3}\hspace{1.5em}
    \textbf{Linning Xu}\authorrefmarkl{1}\authorrefmark{2}\hspace{1.5em}
    \textbf{Dahua Lin}\authorrefmarkl{1,2,4}\authorrefmark{2}
  }
  \authorblockA{
   \authorrefmarkl{1}The Chinese University of Hong Kong\hspace{2em}
   \authorrefmarkl{2}Shanghai AI Laboratory\\
   \authorrefmarkl{3}Joy Future Academy\hspace{2em}
   \authorrefmarkl{4}CPII under InnoHK\\
  }
  \authorblockA{
    \authorrefmark{2}Corresponding author \\
  }
  \authorblockA{
   Project page: \href{https://HealorCai.github.io/Skel-WAM/}{\textbf{\textcolor{teal}{https://HealorCai.github.io/Skel-WAM}}}
  }
  \vspace{-45pt}
}

\begin{document}
\maketitle
\thispagestyle{empty}
\pagestyle{empty}

\begin{strip}
    \centering
    \includegraphics[width=\textwidth]{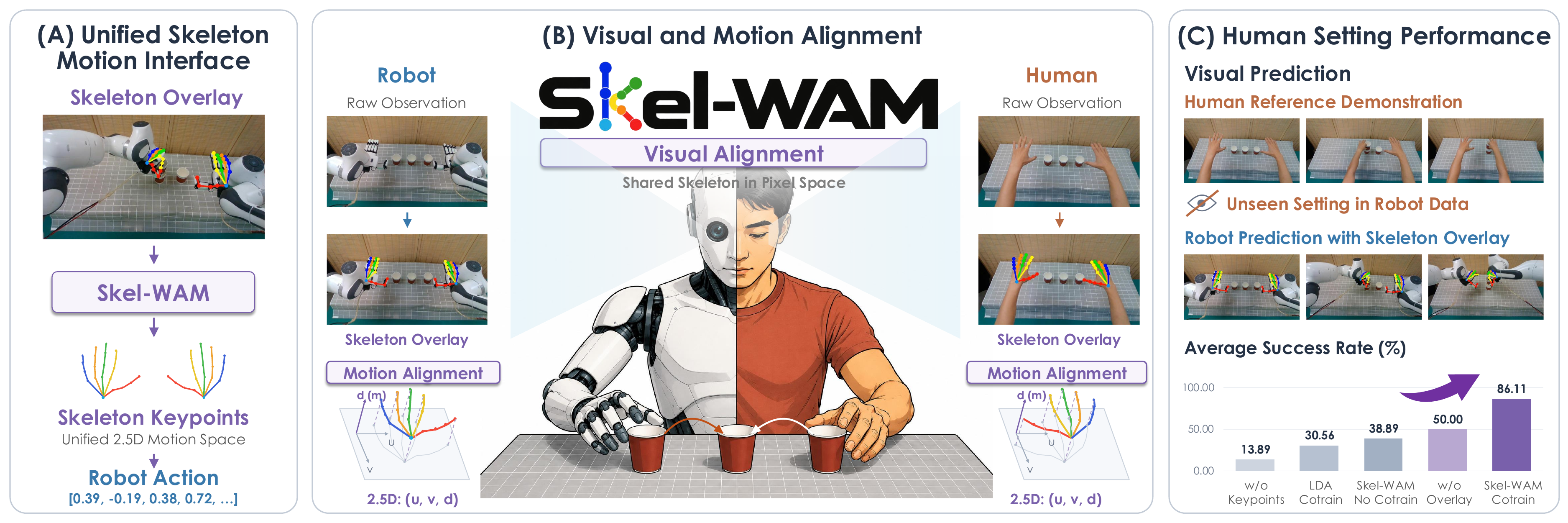}
    \refstepcounter{figure}\label{fig:teaser}
    \vspace*{-10pt}
    \vspace{1mm}
    \parbox{\textwidth}{\footnotesize Fig.~\thefigure.\quad
    \textbf{Skel-WAM} is a world action model that bridges human and robot manipulation through a unified hand-skeleton motion interface.
    Scene-grounded skeleton overlays and structured 2.5-D keypoints represent human and robot motion in a shared space, 
    enabling cross-embodiment alignment and joint prediction of visual dynamics and skeletal motion.
    This representation allows Skel-WAM to learn robot-missing task variations from human demonstrations without robot action labels,
    raising average success from 38.89\% to 86.11\%.
    }
\end{strip}

\begin{abstract}
Robot demonstrations are expensive to collect and often provide limited distributional coverage of task variations. 
Human videos offer a low-cost source of complementary manipulation experience, but learning from them requires bridging embodiment gaps in visual appearance and action spaces.
We introduce Skel-WAM, a world action model that bridges these differences through a unified hand-skeleton motion interface. The key insight is to align human and robot motion through a common hand topology, combining skeleton overlays that ground motion in the scene with structured 2.5-D keypoints that encode explicit hand kinematics. Video and Keypoint Experts jointly learn visual and skeletal dynamics through a Mixture-of-Transformers, while a separate robot-trained Action Expert maps these predictions to executable controls. This separation enables human and robot demonstrations to directly supervise shared dynamics without requiring robot action labels for human videos. Across four real-world bimanual tasks and seven simulated tasks, Skel-WAM achieves average success rates of 79.86\% and 63.29\%, surpassing the strongest baseline by 22.22 and 8.28 percentage points, respectively. Human–robot cotraining more than doubles real-world success on task variations absent from robot training data, from 38.89\% to 86.11\%. These results demonstrate that a shared skeletal interface enables joint learning across human and robot data and expands robot task coverage through complementary human demonstrations.
\end{abstract}

\section{INTRODUCTION}

Scaling robot demonstration datasets has accelerated progress in robot manipulation, but robot data collection remains costly and time-consuming. Moreover, increasing dataset size does not guarantee coverage of task variations.
For a fixed instruction, demonstrations
may capture a nominal task while remaining sparse across object position,
required rotation, environmental disturbance, or manipulation height. A policy
can therefore fit the training distribution and still fail on nearby,
task-consistent cases. Scaling robot datasets improves generalization, but collecting every missing condition for a particular task and workspace remains
costly.

Human videos provide an abundant source of complementary experience. They can
be collected at scale and naturally contain diverse hand-object interactions.
Prior work has used them for video-based planning \cite{unipi} and kinematic
retargeting \cite{telekinesis}. However, converting
this diversity into robot control requires bridging both appearance and action
gaps. Human and robot hands differ in morphology, and human videos lack the
joint positions, end-effector poses, or actuator commands used to supervise
robot policies. Human-specific models such as MANO \cite{mano} provide
kinematic structure but not robot actions. Dense visual motion lacks
finger-level semantics, while direct retargeting relies on
embodiment-specific correspondences.

Alignment alone also leaves the role of human data underspecified. 
Treating humans as another embodiment enables joint learning,
but leaves their concrete contribution to robot policies unclear.
We investigate whether human demonstrations can expand within-task
coverage by supplying variations absent from robot data under the
same instruction. This requires an interface that aligns motion
across embodiments, captures dexterous hand structure, and grounds
that motion in the scene.

We introduce Skel-WAM, a world action model with a unified hand-skeleton motion interface that maps human and robot hands to a common two-hand topology with 21 keypoints per hand. Complementary skeleton overlays ground motion in the scene, while structured 2.5-D keypoints preserve finger identity and explicit trajectories. Video and Keypoint Experts form a Mixture-of-Transformers to learn shared visual and skeletal dynamics from human and robot data. A separate robot-trained Action Expert decodes these predictions into embodiment-specific controls, allowing human videos to supervise shared dynamics without corresponding robot action labels.
We evaluate Skel-WAM on four real-world bimanual tasks and seven simulated
manipulation tasks. It achieves the best average performance among all
evaluated methods. We further hold task language fixed and use human
demonstrations to cover robot-missing variations in position, rotation,
disturbance, and height. Cotraining raises average success on these cases from
38.89\% to 86.11\%, a 47.22-point gain. Furthermore, controlled ablations confirm the
complementary roles of overlays and keypoints. 

Our main contributions are threefold:
\begin{itemize}
    \item a unified hand-skeleton motion interface that aligns human
    and robot motion through visual overlays and structured 2.5-D
    keypoints,
    \item a decoupled Video-Keypoint MoT and robot-only Action
    Expert that learn from human videos without robot action labels,
    and
    \item extensive evaluations in the real world and simulation that show
    consistent gains over strong baselines and a 47.22-point improvement from
    human-covered task variations.
\end{itemize}

\section{RELATED WORK}

\subsection{Human Videos for Robot Learning}

Prior work uses human videos for video or trajectory planning
\cite{unipi,mimicplay,atm,track2act}, retargets estimated human
motion into robot actions \cite{telekinesis,egovla,vitra}, 
or synthesizes robot-format observations by replacing visible
human hands and arms with robot embodiments
\cite{phantom,h2r,egoengine2026,ego2robot2026}.
HandEdit benchmarks dexterous human-to-robot image editing
\cite{handedit2026}. Another line of work treats humans as an
additional embodiment during joint training
\cite{egomimic,egoscale,human_to_robot,egopi2026,egosteer2026}. Other approaches learn from structured human
recordings \cite{humanego2026,aina2025} or use human videos as
in-context task specifications \cite{zerowam2026,host2026}.
Skel-WAM instead studies how human demonstrations can expand
within-task coverage by supplying robot-missing task variations
under fixed task instructions.

\subsection{Cross-Embodiment Motion Representations}

Effective transfer requires a motion representation that remains comparable
across visual and kinematic differences. Prior work uses latent actions, flow,
masks, and sparse tracks or affordances
\cite{lapa,flowwam,maskwam,atm,track2act,motiontracks,pointpolicy,vidbot}.
MANO and retargeting provide explicit human kinematics
\cite{mano,telekinesis}. More closely related methods explicitly model hand structure.
DexWM uses MANO keypoint displacements \cite{dexwm}, Ego-Pi overlays
finger-identified skeletons while mapping human joints into robot
action space \cite{egopi2026}, and OSCAR conditions an omni-embodiment world model
on a common 2D skeleton \cite{oscar2026}.
Skel-WAM instead models human and robot motion in a shared
two-hand skeleton space and jointly predicts future video and
trajectories over 21 keypoints per hand.

\subsection{World Action Models for Manipulation}

A shared representation must also connect predicted futures to control.
ACT and Diffusion Policy predict action chunks directly
\cite{act,diffusionpolicy}, whereas VLAs scale observation-to-action learning
\cite{pi0,groot}. Video policies recover control from predicted futures
\cite{unipi,vpp,gr1,gr2}, and generated videos yield plans through
reconstruction, flow, or inverse dynamics
\cite{dreamgen2025,chen2025lvp,novaflow}. World action models (WAMs)
unify these views by coupling dynamics and control through causal, diffusion,
or latent modeling
\cite{fastwam,causalworldmodel,lda,dreamzero2026,motus}. Recent variants add masks
and 3D targets \cite{maskwam,egowam2026,xwam}, factorized or updatable
simulation \cite{anyworld2026,egosim2026}, and human-video pretraining
\cite{dreamdojo2026}. Skel-WAM jointly predicts video and skeleton with a
Video-Keypoint MoT, then decodes control through a
robot-only action expert. Human demonstrations therefore supervise shared world and motion
dynamics without requiring fabricated robot actions.

\section{METHOD}
\label{sec:method}

\begin{figure*}[t]
    \centering
    \vspace*{1.5mm}
    \includegraphics[width=\textwidth]{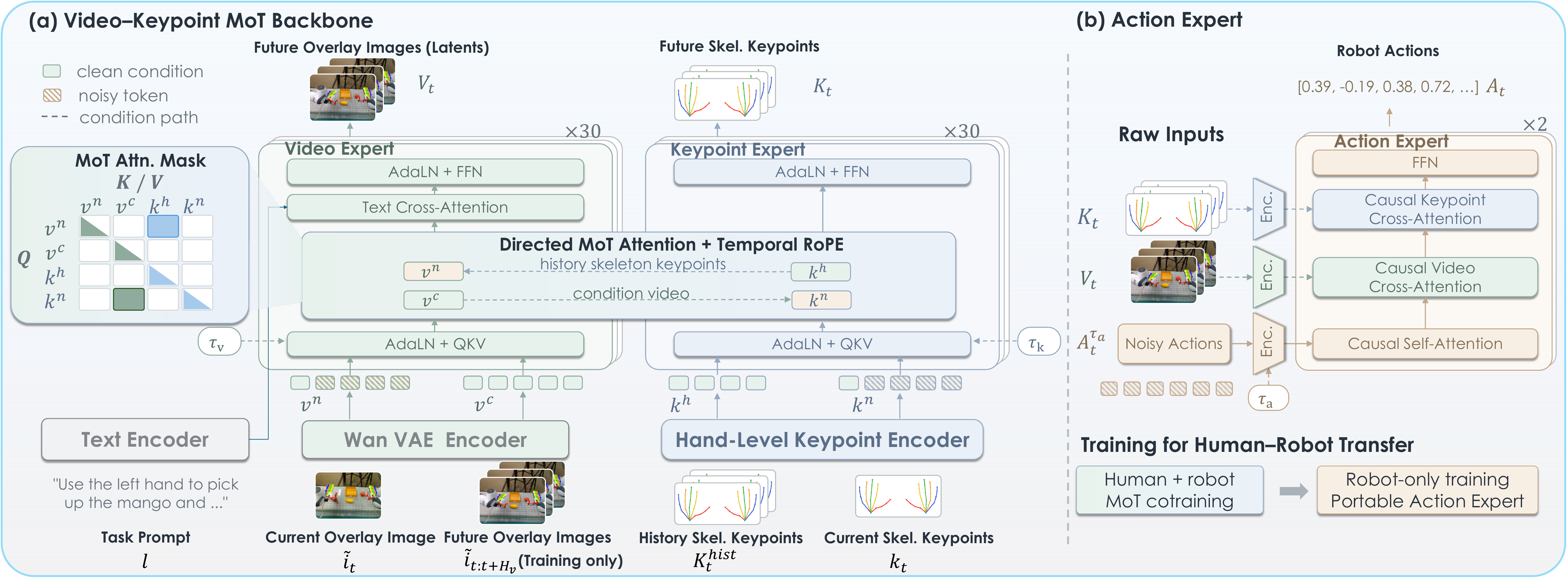}
    \caption{\textbf{Skel-WAM architecture and cotraining.}
    (a) The Video--Keypoint MoT jointly predicts future overlay-video
    latents and structured 2.5-D keypoint trajectories from human and
    robot demonstrations. (b) The causal Action Expert maps these
    predictions to robot actions. Human and robot videos cotrain the
    MoT, while only robot data train the Action Expert, enabling
    learning from human videos without robot action labels.}
    \label{fig:method_overview}
\end{figure*}

In this section, we describe \modelname{} in detail. 
We begin by defining the learning setting with action-labeled robot
demonstrations and human demonstrations without robot action labels
(Sec.~\ref{sec:problem_formulation}).
We then introduce the
unified skeleton motion interface, comprising scene-grounded
overlays and structured 2.5-D keypoints
(Sec.~\ref{sec:skeleton_interface}).
Next, we present the
Video-Keypoint Mixture-of-Transformers for learning shared visual
and skeletal dynamics across human and robot demonstrations
(Sec.~\ref{sec:video_keypoint_mot}).
Finally, we describe
human-robot cotraining of the shared model, robot-only training of
the causal Action Expert (Sec.~\ref{sec:action_cotrain}), and
the flow-matching objectives and implementation details
(Sec.~\ref{sec:implementation_details}).

\subsection{Problem Formulation}
\label{sec:problem_formulation}

At time $t$, the inputs are a task instruction $\ell$, the current RGB image $i_t$, and sampled skeleton keypoint history $K_t^{\mathrm{hist}}=\{k_{t-H_h},\ldots,k_t\}$. Let $v_t$ denote the frozen video VAE latent of the current skeleton-overlay image $\tilde{i}_t$. The model predicts an action chunk $A_t=\{a_t,\ldots,a_{t+H_a}\}$ through intermediate overlay-video latents $V_t=\{v_t,\ldots,v_{t+H_v}\}$ and structured keypoints $K_t=\{k_t,\ldots,k_{t+H_k}\}$, both anchored at the current observation.

Human and robot demonstrations share a skeleton topology but differ in action supervision: $\mathcal{D}_R=\{(I^R,K^R,\ell,A^R)\}$ and $\mathcal{D}_H=\{(I^H,K^H,\ell)\}$. We factor shared prediction as
\begin{equation}
\begin{aligned}
&p_\theta(V_t,K_t\mid v_t,K_t^{\mathrm{hist}},\ell)\\
&\quad=p_\theta^V(V_t\mid v_t,K_t^{\mathrm{hist}},\ell)\,
p_\theta^K(K_t\mid V_t),
\end{aligned}
\label{eq:shared_model}
\end{equation}
and decode robot actions through
$p_\phi(A_t\mid V_t,K_t)$.
Both datasets supervise $p_\theta$, whereas only $\mathcal{D}_R$ supervises
$p_\phi$, separating shared dynamics from robot-specific control.

\subsection{Unified Skeleton Motion Interface}
\label{sec:skeleton_interface}

We represent human and robot hands with a shared topology of two
hands and 21 keypoints per hand. Each skeleton is expressed in two
synchronized forms: a scene-grounded visual overlay and structured
2.5-D keypoints. Together, they preserve hand location, finger
configuration, and bimanual motion without requiring correspondence
between human and robot joint commands.

\begin{figure}[t]
    \centering
    \includegraphics[width=\columnwidth]{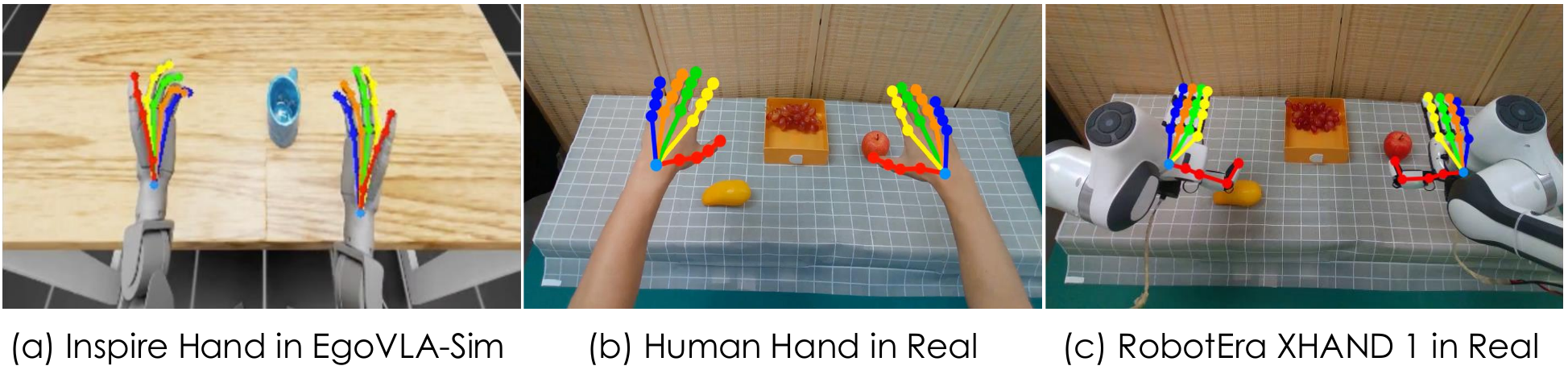}
    \caption{\textbf{Skeleton overlays across embodiments.} (a) Inspire
    dexterous hands in simulation. (b) Human hands and (c) RobotEra XHand1
    dexterous hands in the real world.}
    \label{fig:skeleton_overlay}
\end{figure}

\textbf{Visual skeleton overlay.}
Skel-WAM uses skeleton-overlay images for both observed conditions
and future video prediction. During training, overlays are rendered on
observed and future RGB images. At inference, the Video Expert
directly generates future overlay images.
For human video, WiLoR \cite{wilor} estimates MANO parameters and image-space
keypoints. 
We recover metric keypoints using the measured hand size
and a two-stage camera-space optimization that first estimates wrist
translation and then jointly refines translation and rotation.
For robot videos, we select semantically corresponding URDF links, compute their positions by
forward kinematics, and transform them into the camera frame using calibrated
extrinsics. 
Camera intrinsics project both human and robot keypoints into image coordinates while
retaining metric depth.
We construct $\tilde{i}_t=
\operatorname{Overlay}(i_t,k_t)$ by drawing valid
keypoints and connections on each RGB image
(Fig.~\ref{fig:skeleton_overlay}). Visible elements are rendered in
depth order with depth-adaptive marker sizes, while valid
out-of-frame points remain only in the structured representation.
The overlays thus provide a shared, scene-grounded representation
of human and robot hand motion.

\textbf{Structured 2.5-D keypoints.}
The same skeleton is retained as structured 2.5-D keypoints,
providing an explicit, low-dimensional representation of hand
motion. Each keypoint contains normalized image coordinates
$u=x/W$ and $v=y/H$, together with camera-space depth $d$ in
meters. We standardize these coordinates using fixed statistics
shared across human and robot data, without clipping image
coordinates. Invalid coordinates are zero-filled after
normalization, and validity is encoded as $m=-0.5$ or $+0.5$.
Thus, $k_{t,h}$ contains the coordinates and validity of the 21
keypoints for hand $h$ at time $t$.
Within the Keypoint Expert, the Hand-Level Keypoint Encoder maps
the 21 keypoints of each hand at every frame to a single token:
\begin{equation}
k_{t,h}^{r}=P_k(k_{t,h})
+e_{h}+e_{r},
\label{eq:hand_token}
\end{equation}
where $P_k$ is a learnable linear projection and $r$ identifies the
history condition or noise role. 
Temporal RoPE is applied to encode frame positions.

\subsection{Video-Keypoint Mixture-of-Transformers}
\label{sec:video_keypoint_mot}

We initialize the Video Expert from Wan2.2-TI2V-5B \cite{wan2025wan}. A frozen
Wan VAE encodes skeleton-overlay clips, and a shared patch
projection $P_v$ maps condition and noisy prediction latents into
separate streams. Precomputed T5 embeddings condition the Video
Expert through text cross-attention. Following Fast-WAM
\cite{fastwam}, we initialize transferable Keypoint-Expert weights
by interpolating Video-Expert parameters. Both experts
contain 30 layer-aligned blocks but retain modality-specific
parameters.

The experts separately embed their flow times $\tau_v$ and $\tau_k$
to generate scale, shift, and residual-gate parameters for every
attention and feed-forward branch.
The MoT operates on four token streams shown in
Fig.~\ref{fig:method_overview}(a): video-prediction tokens $v^n$,
video-condition tokens $v^c$, observed keypoint-history tokens
$k^h$, and keypoint-prediction tokens $k^n$.
The current anchors in $v^n$ and $k^n$ remain clean, while their future tokens
receive flow-matching noise. During training, future $v^c$ tokens
encode ground-truth overlay images. At inference, they encode
overlays generated by the Video Expert. The video streams share
$P_v$ and use distinct role embeddings, while the skeleton streams
share $P_k$.
Each MoT block applies a stream-wise attention mask that allows
each query stream to access the following key/value streams:
\begin{equation}
\begin{aligned}
v^{n}&\rightarrow\{v^{n},k^{h}\}, &
v^{c}&\rightarrow v^{c},\\
k^{n}&\rightarrow\{k^{n},v^{c}\}, &
k^{h}&\rightarrow k^{h}.
\end{aligned}
\label{eq:mot_mask}
\end{equation}
Within-stream attention is temporally causal, while the two
cross-modal paths provide full access to their condition streams.
Specifically, $v^{n}$ attends to the skeleton keypoint history
$k^{h}$, and $k^{n}$ attends to the video-condition stream $v^{c}$.
Video and keypoint streams use spatio-temporal and temporal RoPE, respectively, with aligned frame positions. Mixed attention is followed by modality-specific feed-forward layers; only the Video Expert uses text cross-attention. The MoT jointly predicts video-latent and keypoint flow velocities.

\subsection{Action Prediction and Human-Robot Cotraining}
\label{sec:action_cotrain}

The Action Expert in Fig.~\ref{fig:method_overview}(b) 
uses separate encoders to map noisy actions
$A_{t}^{\tau_a}$, conditioning video latents $V_t$, and conditioning
skeleton keypoints $K_t$ into modality-specific tokens. During
training, $V_t$ and $K_t$ are derived from ground-truth robot video
and skeleton sequences, while at inference they are provided by
the MoT predictions.
Each of its two Transformer blocks
sequentially applies causal action self-attention, causal video
cross-attention, causal keypoint cross-attention, and a
feed-forward layer. The causal masks allow each action query to
attend only to action and condition tokens at or before its timestep. 
A final prediction head outputs the action flow velocity.

Human-robot cotraining first optimizes the MoT using overlay
videos and skeleton keypoints from both domains as prediction
targets. We then train the portable robot-specific Action Expert exclusively
on robot data. Its video projection $P_v$ is copied from the
cotrained MoT and kept frozen, and action gradients do not
propagate into the backbone.

\subsection{Flow Matching and Implementation}
\label{sec:implementation_details}

We apply conditional flow matching \cite{flowmatching} to video
latents, skeleton keypoints, and robot actions. For each clean
target $x\in\{V_t,K_t,A_t\}$, we independently sample
$\epsilon_x\sim\mathcal{N}(0,I)$ and a flow time $\tau_x$, and
construct
\begin{equation}
x_{\tau_x}=(1-\tau_x)x+\tau_x\epsilon_x,
\qquad u_x^*=\epsilon_x-x.
\label{eq:flow_target}
\end{equation}
For $V_t$ and $K_t$, only future elements receive flow noise; current anchors remain fixed. Each velocity predictor minimizes
\begin{equation}
\mathcal{L}_x=\mathbb{E}\!\left[
\omega_x(\tau_x)
\frac{\sum_i M_{x,i}(\hat{u}_{x,i}-u_{x,i}^*)^2}
{\max(1,\sum_i M_{x,i})}\right],
\label{eq:masked_flow_loss}
\end{equation}
where $\omega_x$ is the scheduler weight. The masks ignore invalid
keypoint coordinates and invalid action steps, while retaining
validity supervision for all future keypoints. Robot-only joint
training minimizes
$\mathcal{L}_V+2\mathcal{L}_K+2\mathcal{L}_a$.
Human-robot MoT cotraining minimizes
$\mathcal{L}_V+2\mathcal{L}_K$, followed by Action Expert training
with $\mathcal{L}_a$.

We use 1000 flow timesteps with SNR shifts of 5 for video and 1 for
keypoints and actions. With probability 0.5, video-condition latents
receive noise from $[0,0.2]$, while the current anchor remains clean
\cite{causalworldmodel}. Each clip pairs 13 keypoint frames with
$384\times384$ images at real-robot and simulation strides of 6 and 3,
yielding 73- and 37-step action chunks. 
The keypoint history contains 3 and 4 frames for real-robot and simulation data, respectively.
Real-robot control runs at 30~Hz with 42-D actions comprising per-arm 3-D end-effector position,
6-D rotation, and 12 hand joint angles. Simulation uses 38-D joint-angle actions.
Using eight NVIDIA A800 GPUs, robot-only training jointly optimizes
all experts for 80k steps. Human-robot transfer cotrains the MoT for
80k steps at a $2{:}1$ robot-to-human ratio, then trains the Action
Expert on robot data for 50k steps. Inference sequentially samples
$V_t$, $K_t$, and $A_t$ using 3/3/5 steps in the
real world and 10/10/5 in simulation. All flows run from noise to data
with fixed observed anchors, requiring no future observations.

\begin{figure*}[t]
    \centering
    \vspace*{1.5mm}
    \includegraphics[width=0.85\textwidth]{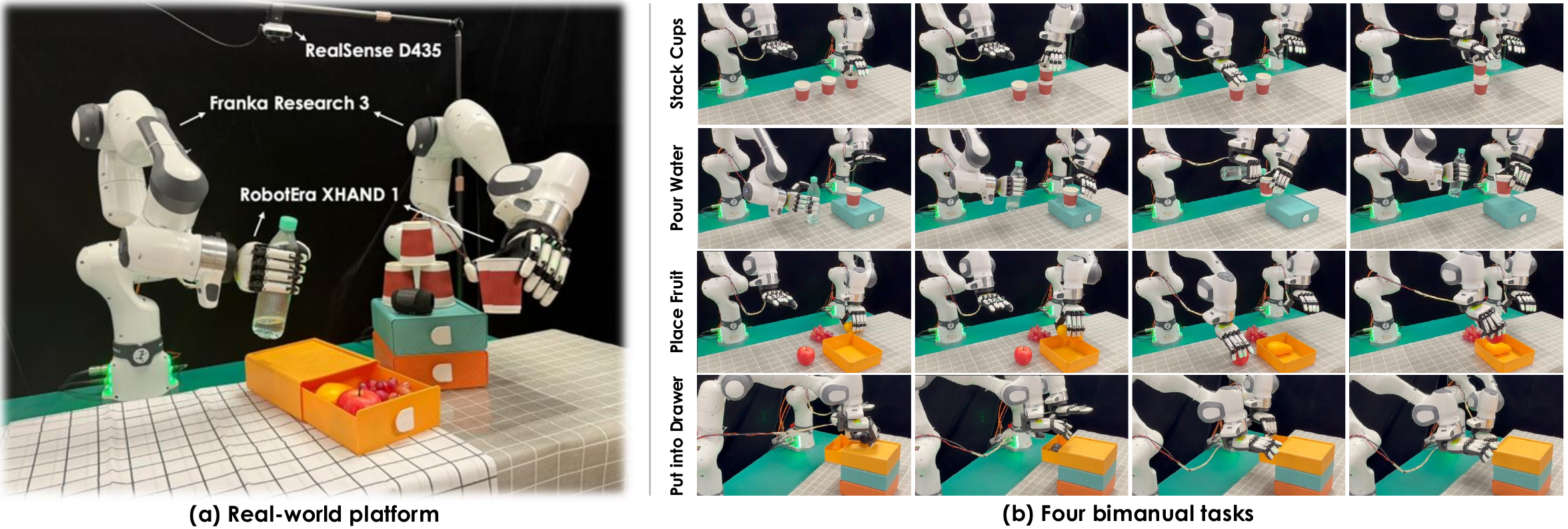}
    \caption{\textbf{Real-world benchmark.} (a) Real-world experimental platform with two
    Franka Research~3 arms, RobotEra XHand1 dexterous hands, and an Intel
    RealSense D435 camera. (b) Execution sequences for Stack Cups, Pour Water,
    Place Fruit into Box, and Put a Wrist Guard into a Drawer.}
    \label{fig:real_benchmark}
\end{figure*}

\begin{table*}[t]
\centering
\caption{\textbf{Real-world results with robot-only training data.} Each entry is SR / PSR (\%).}
\label{tab:real}
\footnotesize
\renewcommand{\arraystretch}{0.9}
\setlength{\tabcolsep}{2pt}
\begin{tabularx}{0.9\textwidth}{l*{5}{Y}}
\toprule
Method & Avg. $\uparrow$ & Stack Cups & Pour Water & Place Fruit & Put into Drawer \\
\midrule
ACT & 12.32 / 44.39 & 11.76 / 58.82 & 16.67 / 60.42 & 0.00 / 37.50 & 20.83 / 20.83 \\
Fast-WAM & 38.89 / 50.69 & 38.89 / 48.61 & 41.67 / 60.42 & 37.50 / 52.08 & 37.50 / 41.67 \\
LDA & 42.01 / 53.30 & 38.89 / 56.94 & 41.67 / 54.17 & 45.83 / 58.33 & 41.67 / 43.75 \\
EgoVLA & 57.64 / 69.10 & 55.56 / 76.39 & 70.83 / 85.42 & 45.83 / 56.25 & 58.33 / 58.33 \\
\rowcolor{gray!15}
\textbf{Skel-WAM} & \textbf{79.86 / 89.24} & \textbf{77.78 / 90.28} & \textbf{87.50 / 95.83} & \textbf{70.83 / 83.33} & \textbf{83.33 / 87.50} \\
\bottomrule
\end{tabularx}
\end{table*}

\section{REAL-WORLD EXPERIMENTS}

We evaluate Skel-WAM on four real-world bimanual tasks using
robot and human demonstrations with complementary task coverage.
We examine four aspects: (1) long-horizon performance
under robot-demonstrated settings, (2) transfer of robot-missing
task variations from human demonstrations, (3) contributions of
skeleton overlays and structured 2.5-D keypoints to human-to-robot
transfer, and (4) generalization to unseen backgrounds.

\subsection{Real-World Experimental Setup}

\textbf{Benchmark.}
Our platform consists of two Franka Research~3 arms equipped with
RobotEra XHand1 dexterous hands and an Intel RealSense D435
ego-view camera (Fig.~\ref{fig:real_benchmark}(a)).
We evaluate four bimanual tasks
(Fig.~\ref{fig:real_benchmark}(b)):
\textbf{\emph{Stack Cups}} sequentially
stacks two side cups onto the center cup. 
\textbf{Pour Water} coordinates a bottle and a cup to perform pouring. 
\textbf{Place Fruit into Box} places a mango and an apple into a box.
\textbf{Put a Wrist Guard into a Drawer} places the wrist guard into a drawer and then closes it.

\begin{figure}[t]
    \centering
    \includegraphics[width=\columnwidth]{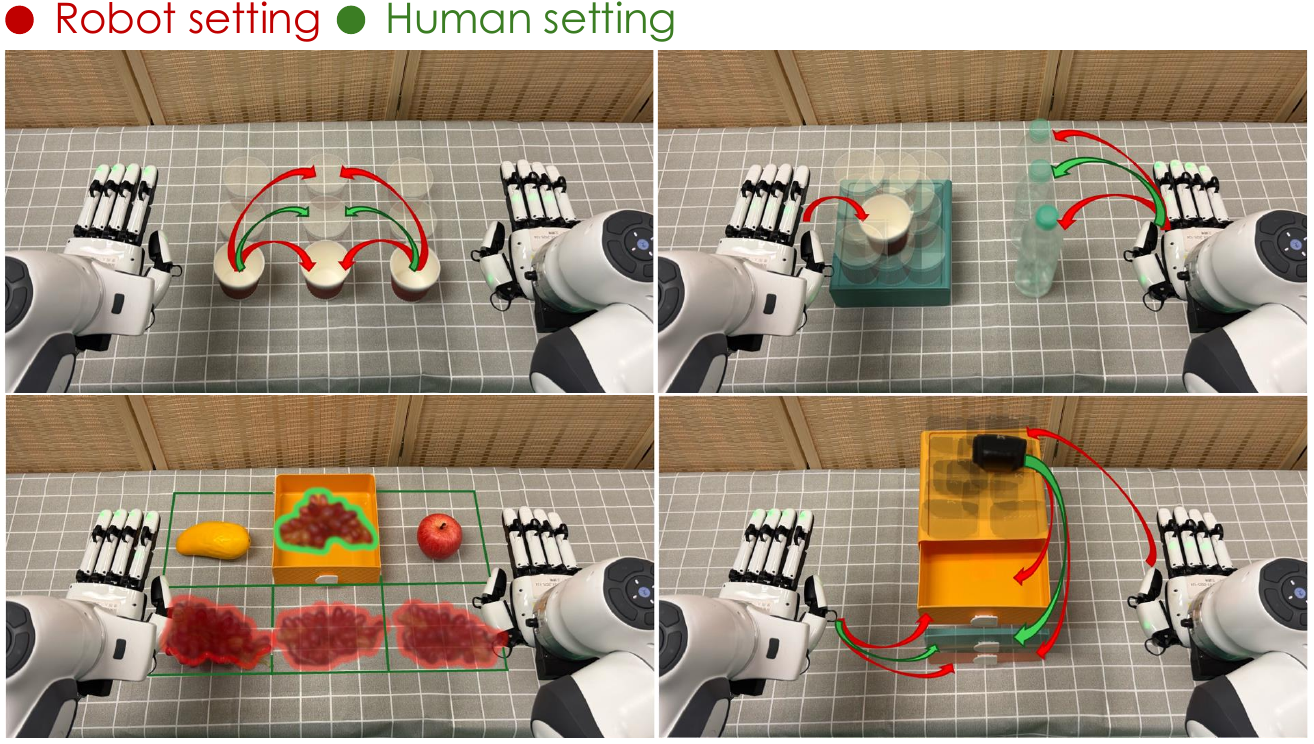}
    \caption{\textbf{Robot and human demonstration settings across the four tasks.}
    Red and green denote robot and human settings, respectively. Human
    demonstrations introduce variations in object position, wrist rotation,
    distractors, and manipulation height.}
    \label{fig:human_settings}
\end{figure}

\textbf{Demonstrations.}
We teleoperate the robot end effectors using relative poses from Vive trackers and control the dexterous hands using MANUS gloves. 
Per task, we collect an average of 50 robot and 25 human demonstrations. Human data cover settings absent from robot demonstrations, varying object position, wrist rotation, distractors, and manipulation height (Fig.~\ref{fig:human_settings}).

\textbf{Baselines.}
We compare with four representative baselines. ACT \cite{act}
directly predicts action chunks from RGB observations, while
Fast-WAM jointly models future videos and robot
actions. LDA~\cite{lda} unifies policy learning, dynamics modeling, and visual
forecasting in the DINO~\cite{dino} latent space. 
EgoVLA~\cite{egovla} is a unified VLA model jointly learning from
egocentric human videos and robot demonstrations.

\textbf{Metrics.}
We evaluate the Robot, Human, and New Background settings using
24, 9, and 6 trials per task, respectively. Stack Cups uses 18
Robot-setting trials to balance the setting distribution. Success
Rate (SR) measures full task completion, while Progress Success Rate
(PSR) measures task-specific milestones. These milestones are
picking and stacking each side cup for Stack Cups, lifting the bottle
and cup for Pour Water, placing each fruit for Place Fruit into Box,
and inserting the wrist guard and closing the drawer for Put a Wrist
Guard into a Drawer.

\begin{table*}[t]
\vspace*{1.5mm}
\centering
\caption{\textbf{Results on robot-missing task variations.}
Cotrained models use robot and human demonstrations, whereas
the no-cotrain model uses robot demonstrations only.
Each entry is SR / PSR (\%).}
\label{tab:human}
\footnotesize
\renewcommand{\arraystretch}{0.9}
\setlength{\tabcolsep}{2pt}
\begin{tabularx}{0.9\textwidth}{l*{5}{Y}}
\toprule
Method & Avg. $\uparrow$ & Position & Rotation & Disturbance & Height \\
\midrule
LDA-cotrain & 30.56 / 49.31 & 33.33 / 41.67 & 22.22 / 44.44 & 44.44 / 61.11 & 22.22 / 50.00 \\
Skel-WAM, no cotrain & 38.89 / 61.11 & 11.11 / 50.00 & 77.78 / 83.33 & 66.67 / 66.67 & 0.00 / 44.44 \\
\rowcolor{gray!15}
\textbf{Skel-WAM, cotrain} & \textbf{86.11 / 95.14} & \textbf{66.67 / 91.67} & \textbf{100.00 / 100.00} & \textbf{77.78 / 88.89} & \textbf{100.00 / 100.00} \\
\midrule
Cotrain w/o both & 16.67 / 31.25 & 11.11 / 19.44 & 33.33 / 61.11 & 22.22 / 33.33 & 0.00 / 11.11 \\
Cotrain w/o keypoints & 13.89 / 34.03 & 11.11 / 25.00 & 22.22 / 55.56 & 22.22 / 38.89 & 0.00 / 16.67 \\
Cotrain w/o overlay & 50.00 / 68.75 & 44.44 / 86.11 & 77.78 / 83.33 & 44.44 / 55.56 & 33.33 / 50.00 \\
\bottomrule
\end{tabularx}
\end{table*}

\subsection{Long-Horizon Performance under the Robot Setting}

Skel-WAM achieves 79.86\% average SR and 89.24\% PSR, exceeding EgoVLA by 22.22 and 20.14 percentage points, respectively (Table~\ref{tab:real}).
Its advantage extends
across all four tasks, which require sequential grasping, bimanual
coordination, distractor-robust localization, and precise placement.
Compared with ACT, which directly predicts actions from RGB
observations, Skel-WAM uses 2.5-D skeleton keypoints to provide
explicit image-space position and depth. This structured motion
interface reduces the grasp drift observed for ACT in Stack Cups.
Fast-WAM predicts visual dynamics but lacks an explicit skeletal
motion representation, leading to mistimed grasps in Stack Cups
and inaccurate cup localization in Pour Water. Skel-WAM instead
combines future visual dynamics with structured hand trajectories,
providing more precise geometric and temporal guidance.
LDA predicts actions through future DINO features. In Place Fruit
into Box, nearby grapes bias its predicted apple-grasp location,
indicating sensitivity to visual distractors. EgoVLA is the
strongest baseline, but its remaining failures also concentrate
on grasp and placement localization. By explicitly grounding hand
motion in image position, depth, and predicted visual context,
Skel-WAM achieves more reliable spatial grounding throughout
long-horizon manipulation.

\begin{figure}[t]
    \centering
    \includegraphics[width=\columnwidth]{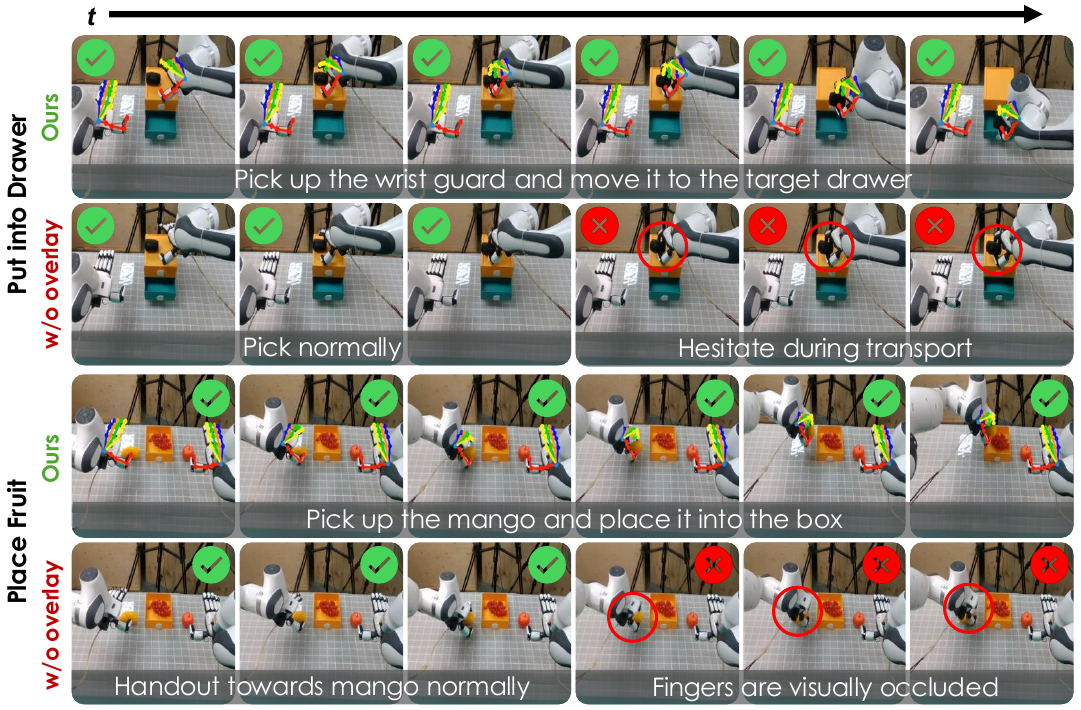}
    \caption{\textbf{Future-video predictions with and without skeleton overlays.} The
    model without overlays stalls while transporting the wrist guard after
    grasping it and loses the hand structure needed to grasp the mango under
    four-finger occlusion in Place Fruit into Box. Full Skel-WAM maintains
    coherent motion and explicit hand structure in both cases.}
    \label{fig:video_generation}
\end{figure}

\begin{figure}[t]
    \centering
    \includegraphics[width=\columnwidth]{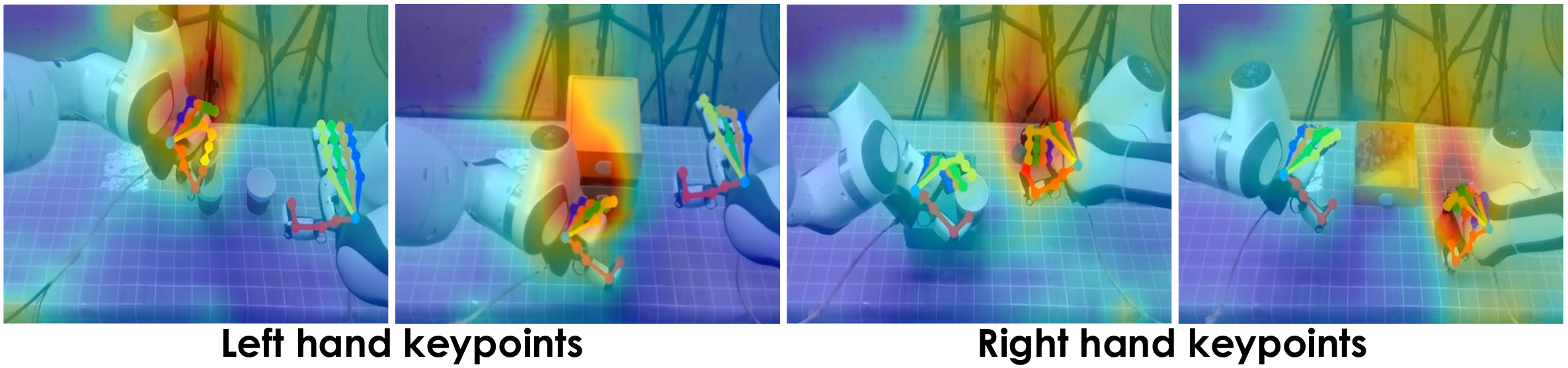}
    \caption{\textbf{Hand-specific keypoint-to-video attention.} Left and right columns show attention from the respective hand keypoint streams, concentrated on corresponding hand regions, indicating spatial alignment with video features.}
    \label{fig:keypoint_video_attention}
\end{figure}

\subsection{Learning Robot-Missing Task Variations from Human Demonstrations}

We evaluate four robot-missing task variations covered only by human demonstrations under fixed task instructions (Fig.~\ref{fig:human_settings}).
They vary middle-cup position in Stack Cups, wrist rotation in Pour Water, distractors in the target box in Place Fruit into Box, and target drawer height.

Human-robot cotraining raises Skel-WAM's average SR from
38.89\% to 86.11\% and PSR from 61.11\% to 95.14\%
(Table~\ref{tab:human}). The gains span all four variations. For
drawer height, cotraining increases SR from 0\% to 100\%,
demonstrating that human videos enable Skel-WAM to acquire task
cases absent from the robot data.
LDA-cotrain directly aligns human wrist poses and MANO parameters
with robot end-effector poses and dexterous-hand actions, yet
achieves only 30.56\% SR and 49.31\% PSR. This suggests that directly bridging embodiment-specific pose and action spaces
retains a substantial transfer gap. Skel-WAM instead maps human and robot motion
into a shared skeleton space, allowing human
videos to supervise shared visual and skeletal dynamics without
robot action labels.

Ablations confirm the complementary roles of skeleton overlays and structured keypoints. Average SR drops from 86.11\% to 50.00\% without overlays and below 17\% without keypoints, regardless of overlays. Keypoints encode finger-level trajectories, while overlays ground them in the scene. Full Skel-WAM maintains coherent wrist-guard transport and hand structure under occlusion during mango grasping; removing overlays causes stalled motion and lost hand structure (Fig.~\ref{fig:video_generation}), supporting the value of both visual and motion alignment for human-to-robot transfer.
Attention maps provide further evidence for this alignment
(Fig.~\ref{fig:keypoint_video_attention}). The left- and
right-hand keypoint tokens attend to their corresponding image
regions, showing that the Keypoint Expert preserves hand identity
and retrieves spatially aligned visual evidence.

\begin{table*}[t]
\vspace*{1.5mm}
\centering
\caption{\textbf{Simulation results.} Each entry is SR / PSR (\%).}
\label{tab:sim}
\scriptsize
\renewcommand{\arraystretch}{0.9}
\setlength{\tabcolsep}{2pt}
\begin{tabularx}{\textwidth}{l*{8}{Y}}
\toprule
Method & Avg. $\uparrow$ & Stack Can & Push Box & Open Drawer & Close Drawer & Flip Mug & Pour Balls & Open Laptop \\
\midrule
ACT & 24.88 / 55.93 & 12.90 / 13.98 & 16.13 / 87.73 & 22.58 / 70.43 & 59.14 / 96.77 & 2.15 / 2.15 & 5.38 / 64.52 & 55.91 / 55.91 \\
LDA & 34.87 / 61.52 & 22.58 / 23.66 & 22.58 / 93.55 & 21.51 / 76.34 & 66.67 / 78.49 & \textbf{50.54 / 50.54} & 2.15 / 43.55 & 58.06 / 64.52 \\
Fast-WAM & 37.33 / 56.22 & 29.03 / 29.03 & 19.35 / 89.25 & 38.71 / 83.33 & 97.85 / 98.92 & 24.73 / 26.88 & 17.20 / 29.57 & 34.41 / 36.56 \\
EgoVLA & 55.01 / 65.31 & 56.99 / 56.99 & 64.52 / 80.65 & 49.46 / 73.16 & 90.32 / 92.83 & 4.41 / 4.41 & \textbf{56.99 / 83.56} & 62.37 / 65.59 \\
\rowcolor{gray!15}
\textbf{Skel-WAM} & \textbf{63.29 / 78.42} & \textbf{58.06 / 59.14} & \textbf{66.67 / 97.85} & \textbf{49.46 / 86.02} & \textbf{100.00 / 100.00} & 45.16 / 45.16 & 27.96 / 61.83 & \textbf{95.70 / 98.92} \\
\bottomrule
\end{tabularx}
\end{table*}

\subsection{Background Generalization}
\begin{figure}[t]
    \centering
    \vspace{-2mm}
    \includegraphics[width=0.8\columnwidth]{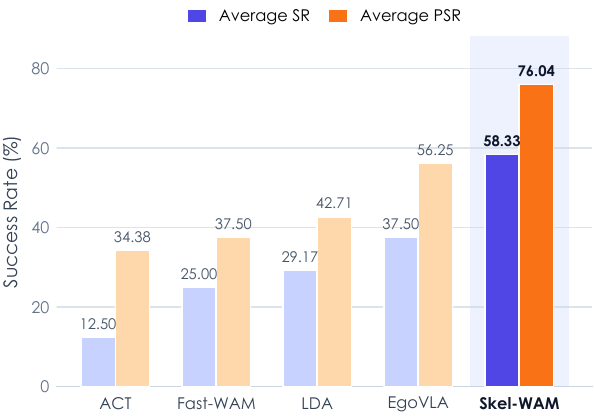}
    \caption{\textbf{Background generalization with robot-only training.}
    Average SR and PSR across four tasks under an unseen background.}
    \label{fig:background_results}
\end{figure}

\begin{figure}[t]
    \centering
    \includegraphics[width=0.9\columnwidth]{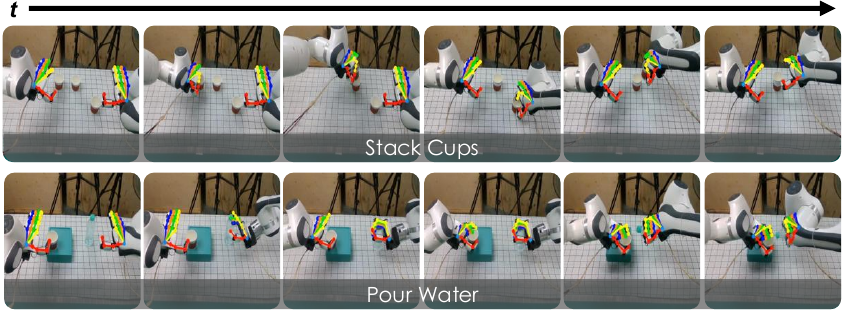}
    \caption{\textbf{Future-video predictions under an unseen background.}
    For Stack Cups and Pour Water, Skel-WAM maintains coherent
    task-relevant motion and preserves the test-time background
    throughout the predicted horizon rather than reverting to the
    training background shown in the inset.}
    \label{fig:background_video_prediction}
\end{figure}

Under the unseen-background setting, Skel-WAM achieves 58.33\%
average SR and 76.04\% PSR, 
surpassing EgoVLA by 20.83 and 19.79 percentage points, respectively (Fig.~\ref{fig:background_results}). 
These gains show that the learned motion representation remains effective
despite changes in scene appearance.
Future-video predictions provide complementary evidence
(Fig.~\ref{fig:background_video_prediction}). Skel-WAM maintains
coherent task-relevant motion while adapting its visual predictions
to the unseen background, as illustrated by the Stack Cups and
Pour Water examples.

\begin{table}[t]
\centering
\caption{\textbf{Simulation ablation.}
Average SR and PSR (\%) over seven tasks are reported.}
\label{tab:sim_ablation}
\begin{tabular}{lcc}
\toprule
Method & SR & PSR \\
\midrule
Skel-WAM & \textbf{63.29} & \textbf{78.42} \\
w/o keypoints & 38.40 & 58.14 \\
w/o overlay & 61.90 & 77.88 \\
w/o both & 34.56 & 55.53 \\
\bottomrule
\end{tabular}
\end{table}

\section{SIMULATION EXPERIMENTS}

We evaluate Skel-WAM on seven EgoVLA-Sim~\cite{egovla} tasks to answer
two questions: (1) How does the unified skeleton motion interface
benefit policy learning from robot demonstrations alone? (2) How
do skeleton keypoints and video latents condition robot action
prediction?

Each task is evaluated over 93 episodes, including 27
in-distribution episodes and 66 episodes with unseen tables or
backgrounds. Object positions are sampled within the benchmark
range in both settings. We compare against the same four baselines
used in the real-world experiments. For fairness, all
models are trained only on robot data, without additional robot pretraining or
human-robot cotraining.

\begin{figure}[t]
    \centering
    \includegraphics[width=0.9\columnwidth]{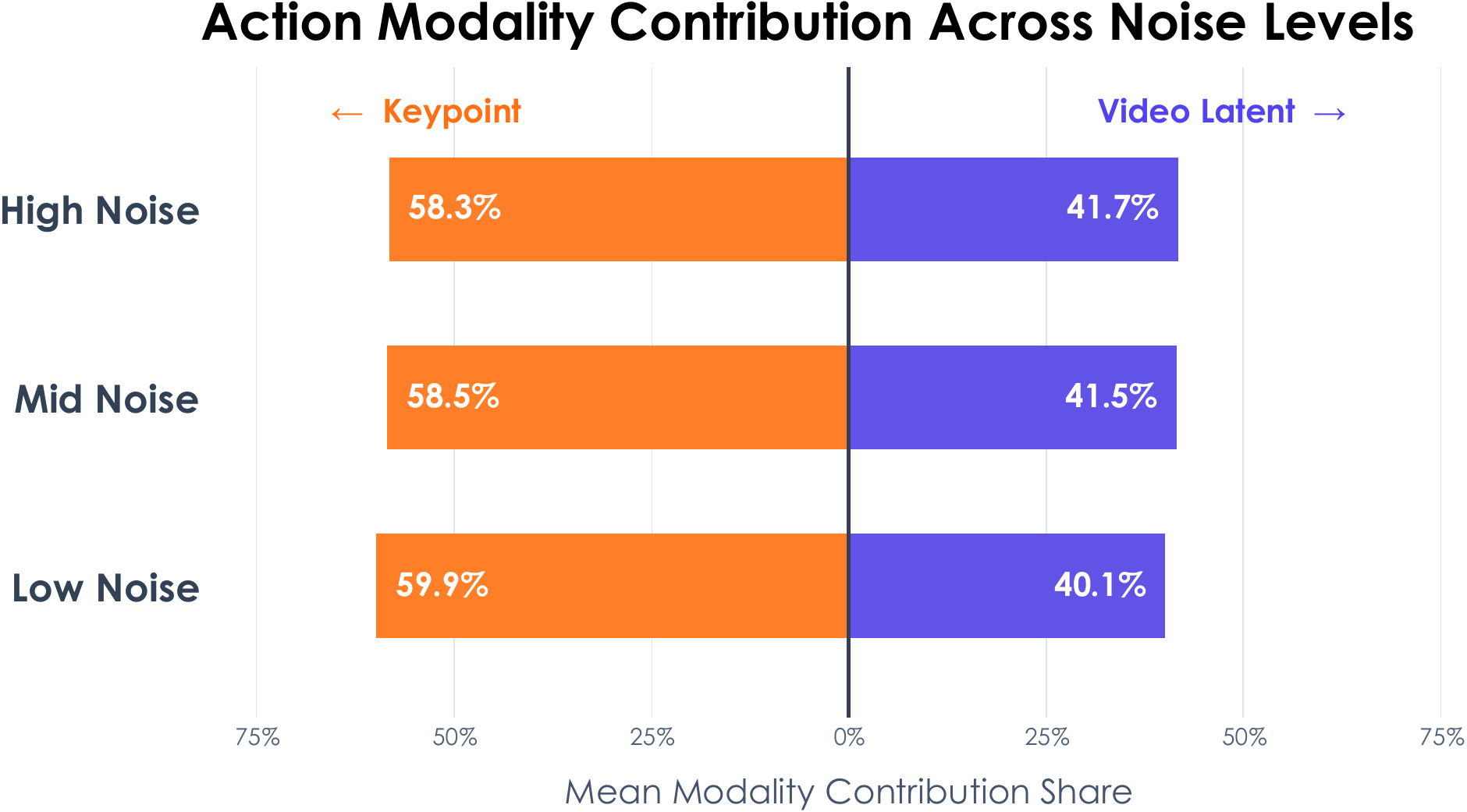}
    \vspace{-4mm}
    \caption{\textbf{Modality residual contribution to action prediction.}
    The gated residual RMS share is averaged across action frames at three
    denoising noise levels. Both conditions contribute throughout denoising,
    with a consistently larger keypoint contribution.}
    \label{fig:action_attention}
\end{figure}

\textbf{Main results and ablation study.}
As shown in Table~\ref{tab:sim}, Skel-WAM achieves 63.29\% SR
and 78.42\% PSR, surpassing EgoVLA by 8.28 and 13.11 percentage
points, respectively. The improvement is observed across most
evaluated tasks. We attribute these gains primarily to structured
keypoint prediction, which provides the Action Expert with explicit
future hand geometry and motion cues complementary to the video
latents. The following ablations support this interpretation.
Table~\ref{tab:sim_ablation} evaluates the two skeleton
representations. Removing structured keypoints decreases average
SR from 63.29\% to 38.40\%, while removing both representations
further reduces it to 34.56\%. In contrast, removing only the
overlay results in 61.90\% SR. Structured keypoints therefore
provide most of the robot-only simulation gain. 
The limited overlay gain is expected as overlays
are most useful during human-robot cotraining, where they reduce the visual
appearance gap.

\textbf{Action modality contribution.}
To quantify video and keypoint conditioning of the Action Expert, let $\Delta_m\in\mathbb{R}^{d_h}$ denote the residual update from modality $m\in\{V,K\}$ after output projection and, for keypoints, the learned gate. We measure its RMS magnitude as $r_m=\lVert\Delta_m\rVert_2/\sqrt{d_h}$ and its normalized share as $s_m=r_m/(r_V+r_K)$, capturing the update magnitude in the action residual stream rather than raw attention weights. Averaged across action frames, keypoints account for 58.3-59.9\% across noise levels and video latents for 40.1-41.7\% (Fig.~\ref{fig:action_attention}). Both modalities contribute to action prediction, with keypoints producing larger residual updates.

\section{CONCLUSIONS}

We introduced Skel-WAM, a world action model for manipulation
that aligns human and robot hand motion through visual skeleton
overlays and structured keypoints. A decoupled Video-Keypoint
MoT learns shared visual and skeletal dynamics, while a robot-only
causal Action Expert maps them to executable controls, enabling
human videos to be used without robot action labels. Skel-WAM
consistently outperforms strong baselines across real-world,
background-shift, and simulation settings. Human-robot cotraining
further raises success from 38.89\% to 86.11\% on four
robot-missing variations under fixed task language. 
These results show that unified hand-skeleton motion transforms
human demonstration diversity into broader coverage of task
variations for robot policies.

\bibliographystyle{IEEEtran}
\bibliography{main}
\end{document}